%% file: neurips_2026.tex
\documentclass{article}

\PassOptionsToPackage{numbers, compress}{natbib}
\usepackage[main, final]{neurips_2026}

\usepackage[utf8]{inputenc} 
\usepackage[T1]{fontenc}    
\usepackage{hyperref}       
\usepackage{url}            
\usepackage{booktabs}       
\usepackage{amsfonts}       
\usepackage{nicefrac}       
\usepackage{microtype}      
\usepackage{xcolor}         
\usepackage{amsmath}
\usepackage{enumitem}
\usepackage{multirow}
\usepackage{graphicx}
\usepackage{booktabs}
\usepackage{makecell}
\usepackage{fontawesome5}
\definecolor{darkblue}{rgb}{0, 0, 0.5}
\hypersetup{colorlinks=true, citecolor=darkblue, linkcolor=darkblue, urlcolor=darkblue}

\title{Rethinking Personalized Generation: Test-Time Alignment via Factorized Ranking Models}

\author{%
  Qiyao Ma
  \quad
  Junshan Zhang\thanks{Equal Advising.}
  \quad
  Zhe Zhao\footnotemark[2]
  \vspace{0.4em}\\
  University of California, Davis\\
  \texttt{\{qiyma,jazh,zao\}@ucdavis.edu}
  \vspace{0.4em}\\
  \faGithub \ \textbf{Code}: \href{https://github.com/Martin-qyma/Rethinking-Personalized-Generation}{https://github.com/Martin-qyma/Rethinking-Personalized-Generation}
}

\begin{document}
\setcounter{footnote}{1}
\maketitle
\setcounter{footnote}{0}
\begin{abstract}
Aligning large language models (LLMs) to diverse user preferences is fundamentally hindered by standard alignment paradigms that optimize for monolithic users. In this work,  empirical studies are first used to reveal the existence of a massive, untapped performance headroom for personalized generation through test-time alignment. We demonstrate that personalized generation is uniquely suited for test-time scaling methods like Best-of-$N$ (BoN) because it can be viewed primarily as a candidate matching problem rather than a generator capability bottleneck. While reward models could in principle exploit this headroom, they are poorly calibrated for personalization, and their billion-parameter scale makes scoring large candidate pools prohibitively expensive. To overcome this limitation, we propose a parameter-efficient framework utilizing million-parameter scale multi-layer perceptron (MLP) ranking models. Our personalized ranking model directly reuses the internal embeddings of the base generator with minimal overhead. By scaling train-time data to provide fine-grained personalized preferences, this million-parameter ranking model accurately scores large candidate pools and can seamlessly guide generation to reduce the cost of materializing $N$ candidates. Extensive experiments on nine datasets spanning three personalized generation settings show that our personalized ranking model effectively exploits the discovered headroom, outperforming billion-parameter generalist reward models on every dataset, with under 0.4\% of their parameters and four orders of magnitude lower scoring latency.
\end{abstract}

\input{intro}
\input{related}
\input{method}
\input{experiments}

\input{conclusion}

\clearpage
\bibliographystyle{plainnat}
\bibliography{ref}
\appendix
\input{appendix}


\clearpage
\input{checklist.tex}

\end{document}

%% file: intro.tex
\section{Introduction}
\label{section:intro}

As large language models (LLMs) are deployed in increasingly subjective settings, from writing in a user's voice to explaining why an item suits them, the assumption behind standard alignment begins to bind. Reinforcement Learning from Human Feedback (RLHF) and its variants~\citep{ouyang2022training, bai2022training} have proven highly effective at instilling general alignment, but they do so under the assumption of a single, monolithic set of human preferences. Human preferences are pluralistic~\citep{sorensen2024roadmap, ryan2025synthesizeme}, and a monolithic reward averages away exactly the idiosyncratic choices that make a response read as if it were written for a particular person~\citep{ma2026personalized, zhao2023group, chen2024pal, poddar2024personalizing}.

\begin{figure*}[t]
    \centering
    \includegraphics[width=\textwidth]{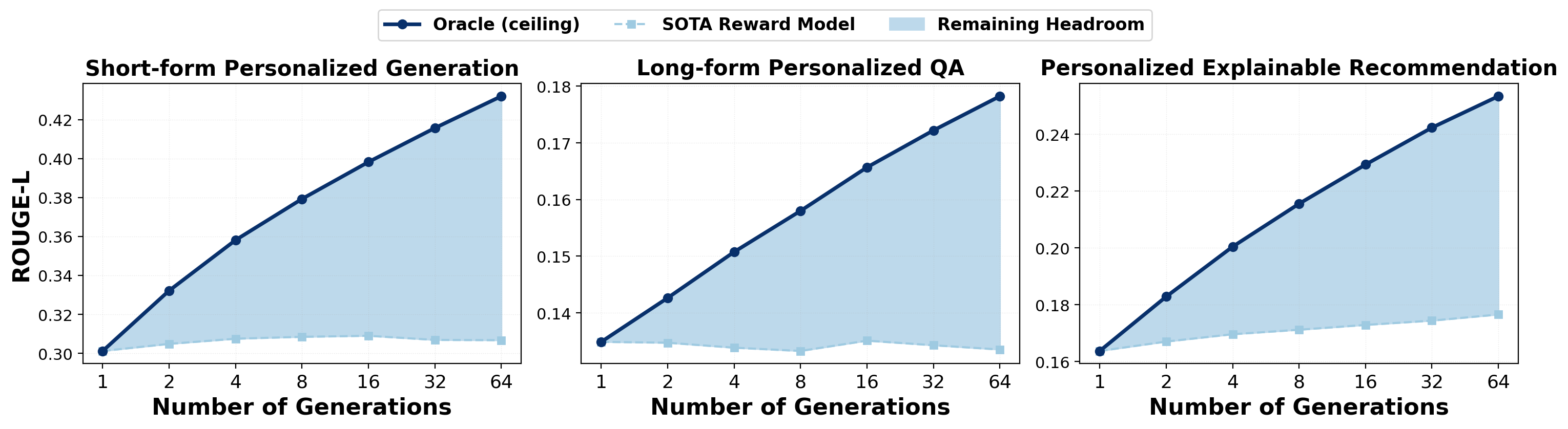}
       \caption{Personalization headroom under Best-of-$N$ sampling. The oracle selects with the evaluation metric itself and upper-bounds any selector; the reward-model line is, at each $N$, the best of the four generalist reward models of Section~\ref{subsec:settings}. The shaded region is the gap between them. Results are averaged over the three datasets of each task.}
    \label{fig:oracle}
    \vskip -0.2in
\end{figure*}

The prevailing response treats this as a capability deficit of the generator and repairs it at training time, by finetuning per user or learning to adapt the generator from user data~\citep{jin2025era, tan2025instant}. We start from a different question: Does the base model already have the capability of producing well-personalized responses that we are failing to pick out? Figure~\ref{fig:oracle} answers with an oracle experiment. When the evaluation metric itself selects the best of $N$ sampled candidates, alignment with the user's own reference keeps improving as the pool grows; in contrast,  when a state-of-the-art reward model selects instead, it plateaus almost immediately and captures only a small fraction of the available gain. The oracle shows that well-matched candidates are already in the pool, whereas naively using the reward model cannot find them: \textbf{the capability is in the generator, and the bottleneck is selection}.

Exploiting this headroom with a standard reward model is costly in two ways. A reward model is a separate billion-parameter transformer, so scoring a pool means re-encoding the same prompt and every candidate from scratch: selection cost grows with $N$, with response length, and with the reward model's width, and in our measurements scoring a pool with an 8B reward model costs about as much as generating it. Beyond cost, generalist reward models are trained to rank responses by universal qualities, and Figure~\ref{fig:oracle} shows this is the wrong yardstick for a pool whose candidates are all fluent and all on topic but differ in whose voice they are written in.

We therefore decouple selection from language understanding. The generator has already read the query, the user profile, and each candidate, and its final-layer hidden states carry that understanding; we read those states directly and train a small multi-layer perceptron (MLP) of 1M to 30M parameters to map them to a scalar score. No text is re-encoded and no per-user parameters are learned. Because the ranker operates in a fixed feature space, it cannot learn a representation of its own; whatever it learns about a user's choices must come from the supervision. We build that supervision from the generator itself. For each training prompt we sample a pool of candidates and label every candidate with its ROUGE-L against the reference the user actually wrote. The pool supplies realistic hard negatives, responses that are well formed but not what this user would have written, and pointwise regression on these labels calibrates the ranker for exactly the selection task it faces at test time. The loss itself is not the point: pairwise and listwise objectives perform comparably.

With selection nearly free, the remaining cost of Best-of-$N$ is the pool itself: every candidate must be fully generated before it can be scored. This raises the question of whether the ranker can act before the pool exists. Because it scores candidates from the generator's own hidden states, it can steer decoding directly: when the generator is uncertain about the next token, we nudge its choice toward what the ranker predicts the user would prefer. This recovers part of the selection gain at the cost of a single generation; most of the headroom still requires scoring a pool, so we present guided decoding as a complement to selection rather than a replacement for it.

We evaluate on nine datasets spanning three personalized generation settings: short-form personalized generation~\citep{salemi2024lamp}, long-form personalized question answering~\citep{salemi2025lamp}, and personalized explainable recommendation~\citep{ma2024xrec}. Our ranker has a few million parameters; the reward models it is compared against have eight billion. Against four generalist reward models, it is nonetheless the best selector on every dataset and keeps improving with the pool size where they plateau or decline. We then finetune the best of those reward models per dataset on the same candidate pools with the same objective, which sets the ceiling a model of that size can reach on this task. The finetuned model is ahead, as it should be, but the ranker recovers most of its gain with a model more than two hundred times smaller, at orders of magnitude lower latency, and without finetuning a billion-parameter model for every dataset. Without any preference labels, it also matches personalized reward models that fit per-user parameters from labeled comparisons, and its selections transfer to a metric it was never trained on.

In summary, our main contributions are summarized as follows:
\begin{itemize}[leftmargin=*]
    \item \textbf{Empirical Discovery of Personalization Headroom:} An oracle Best-of-$N$ analysis shows that the base generator already produces well-personalized candidates while generalist reward models recover only a small fraction of the gain, reframing personalized alignment as a candidate selection problem.
    \item \textbf{Introduce Personalized Ranking Model:} A million-parameter MLP that reuses the generator's own hidden states in place of a billion-parameter reward model. Trained on candidate pools with realistic negatives, it outperforms generalist reward models over two hundred times its size and extends to single-pass guided decoding.
    \item \textbf{Comprehensive Experimental Validation:} Extensive evaluations across nine datasets and three paradigms demonstrate that our million-parameter MLP recovers most of the gain of explicitly finetuned, billion-parameter SOTA reward models at a fraction of their size and inference cost, and matches personalized reward models without per-user preference labels.
\end{itemize}

%% file: related.tex
\section{Related Work}
\label{section:related_work}

\paragraph{Generalized Alignment and Reward Modeling.}
RLHF~\citep{ouyang2022training, bai2022training} and DPO~\citep{rafailov2023direct} align LLMs to a monolithic notion of preference through reward models trained with the Bradley-Terry objective~\citep{10.1093/biomet/39.3-4.324}. This averages out pluralistic tastes, and a Bradley-Terry reward is identified only up to a prompt-dependent shift, so its scores are not calibrated across the large candidate pools that test-time selection must compare~\citep{zhu2023principled}. We instead frame personalization as candidate matching and train a lightweight ranking model on the generator's own embeddings with a pointwise objective. The objective itself is not a contribution: pairwise and listwise alternatives perform comparably under the same architecture and data (Appendix~\ref{sec:loss comparison}).

\paragraph{Personalized Generation.}
Early work injects persona descriptions or retrieved history into the prompt~\citep{zhang2018personalizing, salemi2024lamp} or fine-tunes per user~\citep{jin2025era, tan2025instant}, treating personalization as a generator deficit. A second line personalizes the reward model: PAL~\citep{chen2024pal}, VPL~\citep{poddar2024personalizing}, PReF~\citep{shenfeld2025language}, and LoRe~\citep{bose2025lore} place each user in a low-dimensional preference space that is fit or inferred from labeled pairwise comparisons, while GPO~\citep{zhao2023group} and SynthesizeMe~\citep{ryan2025synthesizeme} adapt an LLM judge from a user's labeled examples. All of them require per-user preference annotations and, as deployed, a separate encoder pass per candidate. Our ranker needs neither: it derives the user signal from the interaction history already present in the generation context and recycles the generator's hidden states. Appendix~\ref{app:personalized_rm} compares against all six on XRec, the setting in which per-user supervision can be constructed.

\paragraph{Test-Time Alignment.}
Scaling inference compute can rival scaling model size~\citep{muennighoff2025s1, zhang2025survey}. Best-of-$N$ sampling and guided decoding~\citep{xu2024genarm, khanov2024args, yang2021fudge, chakraborty2024transfer, li2024cascade} rely on a reward model to select or steer and inherit its per-candidate encoding cost; even the lightest of them, FUDGE's future discriminators, re-encode the partial sequence with their own network rather than reading the generator's states, and all target a single monolithic reward rather than an individual user. UserAlign~\citep{puadurean2026inference} and T-POP~\citep{qu2025t} personalize at inference time but elicit preferences through interactive pairwise queries cast as bandits, whereas our ranker issues no queries and needs no labels beyond the user's existing history. Our ranker brings the selection cost to near zero by scoring the generator's own hidden states, and because it is differentiable in those states it can also steer decoding directly.

%% file: method.tex
\section{Methodology}
\label{section:method}

\begin{figure*}[t]
    \centering
    \includegraphics[width=\linewidth]{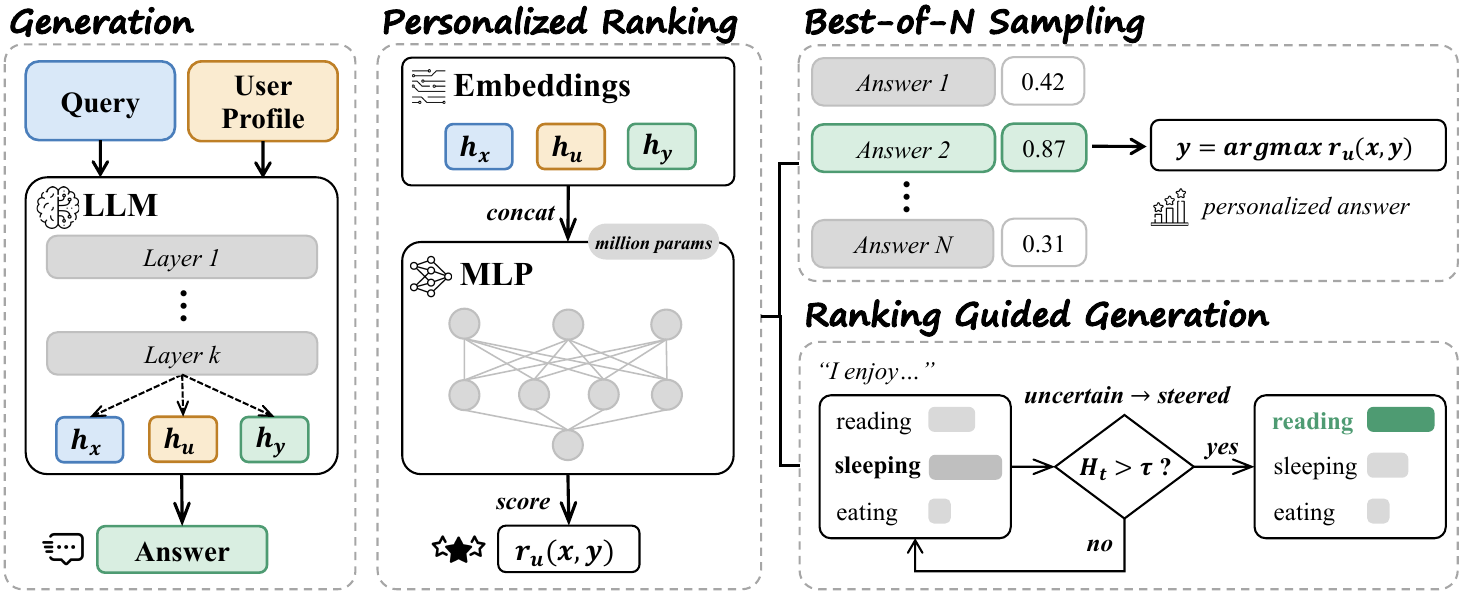}
    \caption{Overview of our decoupled personalized ranking framework. Our personalized ranking model directly recycles the final hidden states of query, user profile, and answer from the LLM generator. This parameter-efficient model is deployed in two inference settings: Best-of-$N$ Sampling and Ranking Guided Generation, where it steers the decoding trajectory when the token distribution entropy ($H_t$) exceeds a predefined threshold ($\tau$).}
    \label{fig:framework}
\end{figure*}

In this section, we formalize the personalized test-time alignment problem and introduce our parameter-efficient, decoupled ranking framework. We begin by defining the core objective of Best-of-$N$ sampling and identifying the severe computational bottlenecks inherent in standard reward models. Subsequently, we detail our lightweight ranking architecture and the preference optimization strategy tailored specifically for learning personal preferences. We conclude with an asymptotic efficiency analysis to theoretically validate the massive scalability of our approach. An overview of our methodology is illustrated in Figure \ref{fig:framework}.

\subsection{Problem Formulation: Personalized Test-time Alignment}
Let $U$ denote a set of users, $X$ the space of input prompts, and $Y$ the space of generated responses. In standard generalized alignment, the goal is to optimize a policy $\pi_\theta(y|x)$ that maximizes a globally shared reward. However, in personalized alignment, the objective shifts to finding an optimal user-specific policy $\pi^*(y|x, u)$ that caters to the idiosyncratic preferences of user $u \in U$.

Instead of actively finetuning the base policy $\pi_\theta$ for every user, test-time alignment employs Best-of-$N$ (BoN) sampling. Given a prompt $x$, the base generator samples a diverse candidate pool of $N$ responses, denoted as $C = \{y_1, y_2, \dots, y_N\}$. The objective is to learn a personalized scoring function, $\hat{r}_u(x, y)$, which identifies the optimal candidate $y^*$ that maximizes the user's implicit utility:
\begin{equation}
    y^* = \arg\max_{y_i \in C} \hat{r}_u(x, y_i)
    \label{eq:bon}
\end{equation}
In standard paradigms, $\hat{r}_u(x, y)$ is parameterized as a massive LLM. Executing Equation~\ref{eq:bon} requires $N$ independent, full forward passes through this heavyweight encoder, creating a severe computational bottleneck that limits the feasible size of $N$ and, consequently, the ability to exploit the personalization headroom.

\subsection{Personalized Ranking Model}
\label{subsec:personalized ranking model}
\paragraph{Decoupling Understanding from Scoring.}
To eradicate the redundancy of secondary encoding, we decouple the language understanding component from the preference scoring component. The base LLM generator has already comprehended the prompt $x$ and synthesized the response $y_i$, so its internal representations already contain dense, high-quality semantic features. Rather than re-encoding the text, our framework directly reads the generator's own final-layer hidden states.

\paragraph{Recycled Embeddings.}
Three texts take part in personalized generation: the task query $x$, the user profile $u$, and the sampled response $y_i$. The profile is the user's interaction history as provided by the benchmark: retrieved history items for LaMP and LaMP-QA, and a pre-written user summary for XRec. For each of the three texts we take the generator's final-layer hidden state at its last token, denoted $h_x, h_u, h_{y_i} \in \mathbb{R}^{d}$, where $d$ is the hidden dimension (the last hidden token in Figure~\ref{fig:framework}). Appendix~\ref{app:impl} specifies the profile text used for each benchmark and how the embeddings are extracted in our implementation.

\paragraph{User Embeddings.}
The user embedding $h_u$ involves no separate user encoder and no per-user learned parameters. It is the generator's own summary of the profile text, so the same frozen model that writes the response also represents the user. Because $h_x$ and $h_u$ depend only on the prompt, they are computed once per query and shared across all $N$ candidates.

\paragraph{Ranking Heads.}
We map the three vectors to a scalar utility through a small MLP $f_\phi$
with $L$ hidden layers:
\begin{equation}
\hat{r}_u(x, y) \;=\; f_\phi\!\bigl(h_x,h_u,h_y\bigr)
\;=\; \mathbf{w}^\top\, g_L \circ g_{L-1} \circ \cdots \circ g_1
       \bigl([h_x \,\Vert\, h_u \,\Vert\, h_y]\bigr),
\label{eq:ranker}
\end{equation}
where $\Vert$ denotes concatenation, each block $g_\ell$ is the activation layer, and $\mathbf{w}$ is the final scalar projection. All configurations are trained from scratch with the same protocol. Because the semantic heavy lifting is offloaded to the LLM generator, the ranking model only needs to learn which candidate in a user-specific pool comes closest to what that user would have written. This is a regression problem in a fixed feature space rather than a representation-learning problem.

\subsection{Pointwise Preference Optimization}
\label{subsec:loss function}
\paragraph{Objective}
The ranking model $f_\phi$ is agnostic to the training objective: any differentiable ranking loss can be applied to the recycled embeddings of Section~\ref{subsec:personalized ranking model}. We adopt pointwise regression by default. Let $s^*(x, y_i, u) \in \mathbb{R}$ denote the target utility of candidate $y_i$ for user $u$. We minimize
\begin{equation}
    \mathcal{L}_{\text{MSE}}(\phi) = \mathbb{E}_{(x, y_i, u) \sim \mathcal{D}} \left[ \left( f_\phi(h_x, h_u, h_{y_i}) - s^*(x, y_i, u) \right)^2 \right]
\end{equation}
which maps every candidate onto a shared, absolute utility scale.

\paragraph{Target Standardization.}
In practice we standardize $s^*$ within each candidate pool, to zero mean and unit variance over the $N$ candidates of a prompt. Best-of-$N$ selection is invariant to per-prompt shifts, so this focuses the regression on within-pool discrimination rather than on predicting each prompt's overall difficulty.

\paragraph{Why Pointwise?}
Two properties of our setting favor pointwise regression. First, our supervision is cardinal rather than ordinal: each candidate carries an absolute target, whereas pairwise or listwise objectives are invariant to monotone transformations of the targets and therefore discard how much better one candidate is than another. Second, Best-of-$N$ selection and the guided decoding of Section~\ref{sec:ranking_guided} consume absolute utilities that must be comparable across large pools, whereas pairwise scores are identified only up to prompt-dependent shifts. Pointwise training also scales as $\mathcal{O}(N)$ per prompt rather than $\mathcal{O}(N^2)$ for all-pairs objectives.

\paragraph{Alternative Objectives.}
Alternatives include the pairwise Bradley-Terry objective standard in reward modeling~\citep{10.1093/biomet/39.3-4.324} and RankNet~\citep{10.1145/1102351.1102363}, and listwise objectives such as ListNet~\citep{10.1145/1273496.1273513}, ListMLE~\citep{10.1145/1390156.1390306}, and LambdaRank~\citep{burges2010from}. Appendix~\ref{sec:loss comparison} compares these objectives under identical architecture, data, and protocol. All objectives perform comparably, so pointwise MSE is a well-founded default rather than a requirement, and the gains of our framework stem from embedding recycling and train-time candidate scaling rather than from the choice of loss.

\paragraph{Train-Time Data Scaling.}
To ensure this minimal architecture achieves high discriminative power, data construction is paramount. Rather than utilizing randomly sampled negative responses, we populate the training distribution $\mathcal{D}$ with candidates directly generated by the base LLM generator $\pi_\theta$. The target utility $s^*(x, y_i, u)$ is the downstream evaluation metric of $y_i$ against the reference authored by user $u$ (ROUGE-L in our experiments). This strategy yields hard negatives: responses that are structurally sound and factually correct, but stylistically or semantically misaligned with the specific user $u$, and therefore receive low targets. Training the ranking model to minimize the MSE over these hard negatives bridges the representation gap typically associated with low-parameter models.

\subsection{Ranking Guided Generation}
\label{sec:ranking_guided}

Best-of-$N$ sampling pays for $N$ full decodings. We instead use the trained personalized ranking model as a single-pass decoding-time signal: the LLM is run once, and the ranking model only intervenes when the model is genuinely uncertain about the next token.

\paragraph{Confidence Gate.}
Let $\boldsymbol{\ell}_t \in \mathbb{R}^V$ denote the next-token logits produced by the LLM generator at decoding step $t$, where $V$ is the vocabulary size. We measure the LLM generator's uncertainty at step $t$ by the next-token entropy
\begin{equation}
  H_t \;=\; -\sum_{v \in \mathcal{V}} p_t(v)\,\log p_t(v),
  \qquad
  p_t \;=\; \mathrm{softmax}(\boldsymbol{\ell}_t).
\end{equation}
With threshold $\tau$, we leave confident steps unchanged:
\begin{equation}
  \text{if } H_t \le \tau:\quad
  y_t \;=\; \arg\max_v\,\boldsymbol{\ell}_t.
\end{equation}

\paragraph{Steering on Uncertain Tokens.}
When $H_t > \tau$, we ask the ranking model which direction in hidden space would increase its predicted reward, and project that direction onto the vocabulary through the LLM head. Concretely, let $W \in \mathbb{R}^{V \times d}$ denote the LLM head weight matrix, where $V$ is the vocabulary size and $d$ is the hidden dimension. We then compute:
\begin{equation}
  \hat{r}_u(x, y) \;=\; f_\phi\!\bigl(h_x,h_u,h_y\bigr),
  \qquad
  g_t \;=\; \nabla_{\mathbf{h}_t}\,\hat{r}_u(x, y) \;\in\; \mathbb{R}^{d},
  \qquad
  \mathbf{u}_t \;=\; W\,g_t \;\in\; \mathbb{R}^{V},
\end{equation}
and emit the argmax of the blended logits
\begin{equation}
  \boldsymbol{\ell}'_t \;=\; \boldsymbol{\ell}_t \;+\; \alpha_t\,\mathbf{u}_t,
  \qquad
  y_t \;=\; \arg\max_v\,\boldsymbol{\ell}'_t.
  \label{eq:blend}
\end{equation}
The step size is entropy-modulated so that the steering is gentle on
borderline-uncertain tokens and stronger on near-uniform distributions:
\begin{equation}
  \alpha_t \;=\; \alpha_0 \cdot
                 \mathrm{clip}\!\left(\frac{H_t - \tau}{H_{\max} - \tau},\,
                                       0,\,1\right).
\end{equation}
The chosen token is appended to the sequence and decoding proceeds to the next step. Appendix~\ref{app:rgg} reports how $\tau$, $H_{\max}$, and $\alpha_0$ were chosen and how sensitive the results are to them.

\subsection{Time Complexity Analysis}
The primary advantage of our decoupled architecture lies in its asymptotic test-time efficiency.

\paragraph{Reward Model Selection.}
Consider a standard Best-of-$N$ scenario where candidate responses have an average sequence length of $L$. For a standard heavyweight reward model, selecting the best candidate requires passing the prompt and response through the full transformer stack $N$ times. The time complexity for the selection phase scales as $\mathcal{O}(N \cdot L \cdot d_r^2)$, where $d_r$ is the hidden dimension of the reward model. As $N$ grows to exploit the personalization headroom, this linear scaling with respect to sequence length and full model dimensionality becomes prohibitively expensive.

\paragraph{Embedding-Recycling Selection.}
In contrast, our proposed framework operates with an $\mathcal{O}(1)$ amortized encoding cost. The generation of the $N$ embeddings occurs naturally during the candidate generation phase. The subsequent ranking step merely requires a forward pass through the shallow MLP for each candidate. The time complexity for selection is thus reduced to $\mathcal{O}(N \cdot d^2)$, completely independent of the sequence length $L$, where $d$ is the hidden dimension of the LLM generator. Because $d^2 \ll L \cdot d_{r}^2$, the latency overhead introduced by our ranker is virtually negligible. This theoretical efficiency allows our method to scale to large candidate pools in real-time deployments with negligible computational cost; Appendix~\ref{app:cost} reports measured latency and memory.

%% file: experiments.tex
\section{Experiments}
\label{section:experiments}
\subsection{Experimental Settings}
\label{subsec:settings}
\paragraph{Generation and Evaluation.}
We use \textit{Qwen2.5-7B-Instruct}~\citep{qwen2025qwen25technicalreport} as the generative backbone. Following the benchmark protocols, we evaluate with ROUGE-L~\citep{lin2004rouge} against the gold reference, and use ROUGE-1 as a check where noted. In all nine datasets the reference is authored by the target user, so lexical overlap with it measures alignment with that user's revealed stylistic choices rather than generic quality; semantic metrics such as BERTScore~\citep{zhang2019bertscore} abstract away precisely the surface-level nuances in which personalization manifests. Because the ranker is trained on ROUGE-L targets, we additionally report BLEU~\citep{post2018call} in Section~\ref{subsec:more analysis} as a held-out metric, verifying that its gains reflect learned user preference rather than regression to the training metric.

\paragraph{Ranking Baselines.}
We evaluate our approach using several state-of-the-art discriminative reward models (RMs) as ranking baselines:
\begin{itemize}[leftmargin=*]
    \item \texttt{Skywork-Reward-V2-Llama-3.1-8B}~\citep{liu2024skywork}: A data-centric RM trained on the Skywork-Reward Preference 80K dataset, which focuses on high-quality curated pairs and utilizes a robust vanilla Bradley-Terry (BT) loss to maximize reward differences.
    \item \texttt{internlm2-7b-reward}~\citep{cai2024internlm2}: A model trained on a large-scale corpus of 2.4 million preference samples (human and AI-synthesized), providing comprehensive coverage across diverse areas including dialogue, coding, and mathematics.
    \item \texttt{URM-LLaMa-3.1-8B}~\citep{lou2024uncertainty}: An uncertainty-aware reward model that implements attribute-specific value heads to predict the parameters of a normal distribution, allowing for the estimation of reward reliability.
    \item \texttt{ArmoRM-Llama3-8B-v0.1}~\citep{wang2024interpretable}: An Absolute-Rating Multi-Objective RM that employs a Mixture-of-Experts (MoE) gating mechanism to scalarize 19 distinct reward dimensions (e.g., helpfulness, correctness, and verbosity) into a single preference score.
\end{itemize}
Every reward model scores exactly the input our ranker sees: the generation prompt, which carries the query and the user's retrieved profile, together with the candidate response. Personalization signal is therefore available to these models in context, and what we test is whether generically trained preference models can exploit it when ranking large candidate pools. Section~\ref{subsec:more analysis} adds a second, stronger tier in which the best of them is finetuned on our personalized training distribution.


\begin{figure*}[t]
    \centering
    \includegraphics[width=\linewidth]{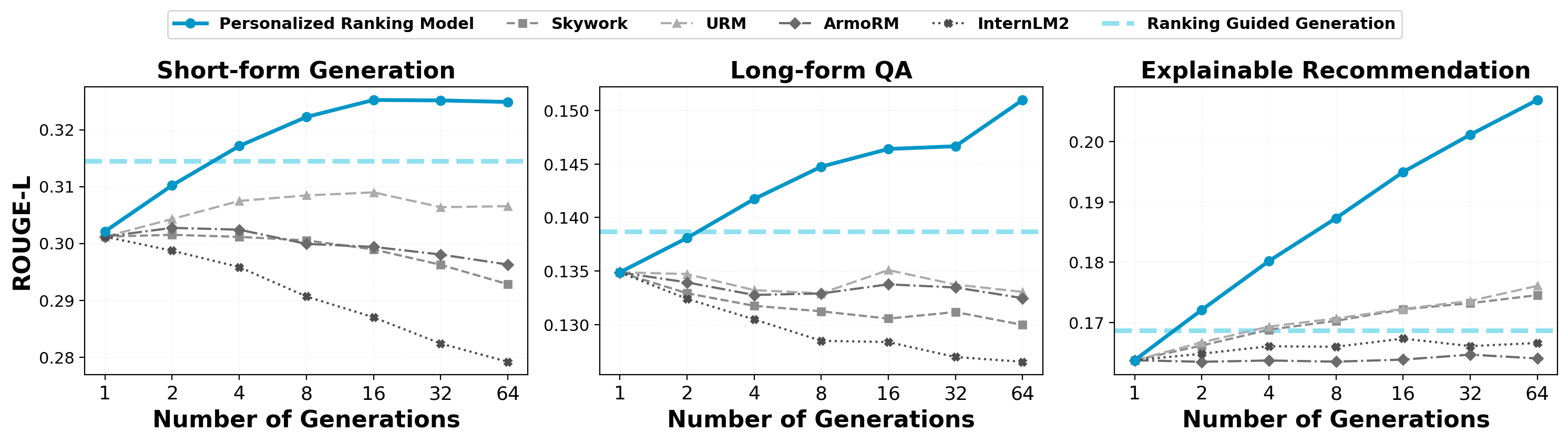}
    \caption{Best-of-$N$ selection with our personalized ranking model and each generalist reward model, averaged over the three datasets of each task. Dashed line: ranking guided generation.}
    \label{fig:ranking}
\end{figure*}

\paragraph{Datasets.}
Our evaluation spans nine datasets across three distinct personalization categories:
\begin{enumerate}[leftmargin=*]
    \item \textbf{Short-form Personalized Generation (LaMP)}~\citep{salemi2024lamp}: We focus on three generative tasks requiring models to adapt to user historical profiles: personalized news headline generation (News), scholarly title generation (Scholarly), and tweet paraphrasing (Tweet).
    \item \textbf{Long-form Personalized QA (LaMP-QA)}~\citep{salemi2025lamp}: This benchmark evaluates informational seeking tasks across three categories: Arts \& Entertainment, Lifestyle \& Personal Development, and Society \& Culture, using personalized rubrics extracted from user narratives.
    \item \textbf{Personalized Explainable Recommendation (XRec)}~\citep{ma2024xrec}: We utilize three domain-specific datasets: Amazon, Yelp and Google, to generate natural language explanations for user-item interactions based on collaborative signals and user profiles.
\end{enumerate}
The size of the user profiles shipped with these benchmarks varies by more than an order of magnitude, from a median of eight interactions per user on XRec to a median of 143 article--headline pairs on News; following each benchmark's protocol, only a small retrieved subset (three items for LaMP, six for LaMP-QA) or a pre-written summary (XRec) enters the prompt. Appendix~\ref{app:datasets} gives self-contained task descriptions, and Table~\ref{tab:history_stats} the per-dataset profile statistics.

\paragraph{Implementation Details.}
The personalized ranking model $f_\phi$ is an MLP over the concatenated $(h_x, h_u, h_y)$ with GELU activations and dropout 0.1. Width-depth pairs $(H, L) \in \{(96,3), (256,3), (1024,3), (2048,4)\}$ give the $\{1.0, 2.8, 12.1, 30.4\}$M-parameter sizes used in our scaling analysis, which we round to 1M, 3M, 10M, and 30M; the 2.8M model is the default elsewhere. We train with AdamW~\citep{loshchilov2017decoupled} at learning rate $5\times10^{-4}$ and weight decay 0.01 for 15 epochs at batch size 64, with gradient clipping at 1.0 and a single seed; the same trainer produces every ranker size and every appendix ablation (Appendix~\ref{app:training}). Each prompt contributes 64 on-policy candidates sampled at temperature 1.0, labeled with ROUGE-L against the gold reference. At inference, ranking guided generation uses $\tau=1$, $H_{\max}=8$, $\alpha_0=5$, and a 3-token warmup; these values were fixed a priori and shared across all nine datasets. Appendix~\ref{app:impl} provides further training details, the sensitivity analysis of the decoding hyperparameters, and measured inference costs. Appendix~\ref{app:full_results} expands every aggregated figure below to all nine datasets, and Appendix~\ref{app:analyses} collects the additional analyses referenced in this section.

\subsection{Main Results}
\label{subsec:main_results}
\label{sec:headroom_analysis}

\begin{figure*}[t]
    \centering
    \includegraphics[width=\textwidth]{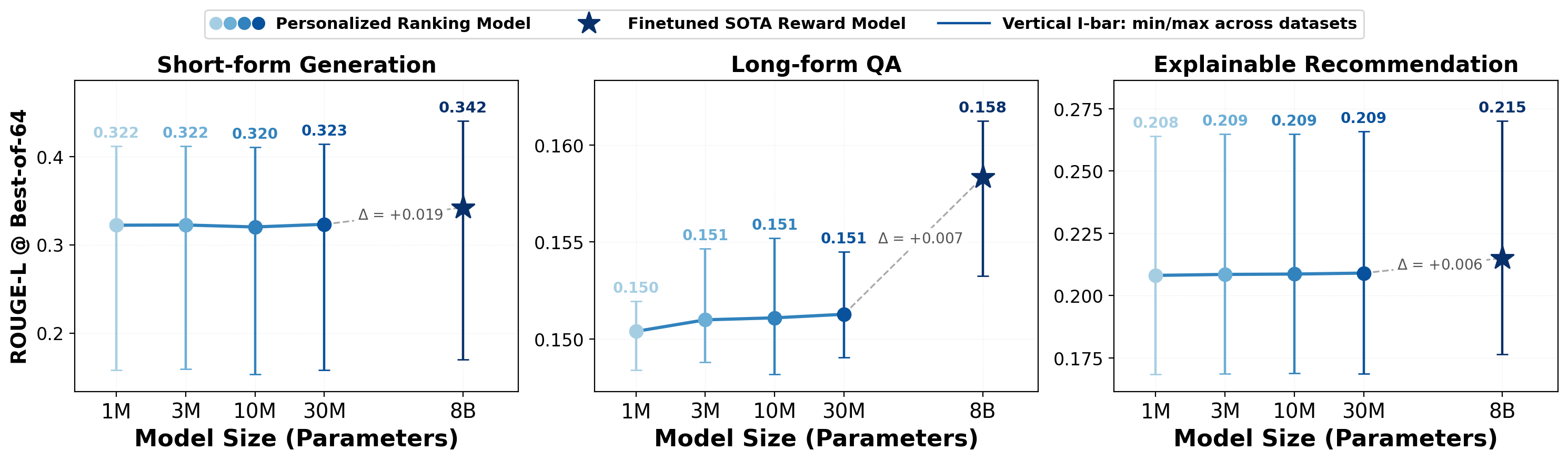}
    \caption{Personalized ranking model size against a finetuned reward model at Best-of-64. Points are means over the three datasets of each task, bars span the per-dataset minimum and maximum.}
    \label{fig:scale}
\end{figure*}

\paragraph{Personalization Headroom.}
Oracle Best-of-$N$ selection picks the candidate with the evaluation metric itself and therefore upper-bounds any selector. Against it we place, at each $N$, the best of the four generalist reward models of Section~\ref{subsec:settings}. Figure~\ref{fig:oracle} shows both averaged per task; Appendix~\ref{appendix:headroom_analysis} gives all nine datasets (Figure~\ref{fig:headroom_full}). The oracle rises steadily up to $N=64$ on every task, with no sign of saturation, so the generator keeps producing better-matched candidates as the pool grows. The best reward model does not follow. It plateaus by $N=16$ on short-form generation and never beats its $N=1$ score on long-form QA. At $N=64$ it captures at most 23\% of the headroom, and under 12\% on eight of the nine datasets. Because one curve keeps climbing while the other stays flat, the gap between them widens with $N$, and every additional sample adds headroom that a generalist selector fails to exploit. This has three implications. First, much of the personalization we seek does not need to be trained into the generator. A frozen model conditioned on the user's profile already produces well-matched responses, and what is missing is the ability to recognize them. Second, test-time compute becomes a lever for personalization, but only if the selector's accuracy scales with the pool. That requires a selector calibrated on the fine, user-specific distinctions among candidates that are all fluent and on topic, rather than on the generic quality differences that generalist reward models are trained to detect. Third, exploiting this headroom requires scoring large pools, so the cost of scoring each candidate becomes the limiting factor. This is what motivates a selector that is both personalized and nearly free to run. The capability is already in the generator; the bottleneck is selection.

\paragraph{Best-of-$N$ Selection.}
Figure~\ref{fig:ranking} compares selectors as $N$ grows from 1 to 64 (per dataset and per reward model in Figure~\ref{fig:rank_full}, Appendix~\ref{appendix:baseline_comparisons}). The two families of selectors follow opposite trends. The generalist reward models stay flat or decline as the pool grows, improving only slowly on explainable recommendation. For them, a larger pool is not an opportunity but a risk. Because they reward generic qualities rather than fit to the user, each added candidate is one more chance to select a response that looks good in general but is wrong for this user. Our ranker instead improves steadily with $N$ on every task, and its advantage over the reward models widens as the pool grows. The first takeaway is that whether test-time compute helps personalization depends entirely on the selector: the same pools yield steady gains under a calibrated ranker and none under a generalist one. The shape of the curve also varies with the task. Gains saturate early on short-form generation, where candidates of a few words differ little from one another, but keep rising on long-form QA and explainable recommendation, where longer responses leave more room for variation. The second takeaway is that pool size should be matched to the task, with small pools sufficient for short outputs and larger ones worthwhile for open-ended responses. We attribute the difference to training on the generator's own candidate pools, which teaches the ranker the fine-grained distinctions that selection actually requires.

\paragraph{Ranking Guided Generation.}
The dashed line marks ranking guided generation, which decodes one response per prompt while the ranker steers uncertain tokens. It reaches the level of Best-of-2 to Best-of-4 of the sampled pool on short-form generation and long-form QA, and of Best-of-1 to Best-of-2 on explainable recommendation, for the cost of a single generation; per dataset it exceeds Best-of-64 only on Tweet and falls below Best-of-1 on News (Figure~\ref{fig:rank_full}). Steering a single trajectory thus recovers a modest part of the selection gain without materializing candidates, while most of the headroom still requires scoring a pool. We conduct a sensitivity analysis by varying each decoding hyperparameter around its default (Appendix~\ref{app:rgg}). ROUGE-L changes by at most about 0.005 on every dataset, so the shared defaults need no per-dataset tuning. Only $\tau$ has a visible effect, and it controls how often the ranker intervenes rather than the final score.

\subsection{More Analysis}
\label{subsec:more analysis}
\paragraph{Scaling Effect and Parameter Efficiency.}
We next ask how much ranker capacity is needed and how the ranker compares with a reward model finetuned for the task. Figure~\ref{fig:scale} varies the ranker from 1M to 30M parameters and adds \textit{Skywork-Reward-V2-Llama-3.1-8B} finetuned on the same training pools with the same pointwise objective (Appendix~\ref{app:training}); points are means over the three datasets of each task with bars spanning the per-dataset range, and Appendix~\ref{appendix:scaling_analysis} gives each dataset (Figure~\ref{fig:scale_full}). Appendix~\ref{app:personalized_rm} further compares against personalized reward models that fit user-specific parameters from per-user preference labels.

Ranker size barely matters: from 1M to 30M parameters the Best-of-64 score moves by at most 0.006 ROUGE-L on any dataset, because the semantic work is done by the generator and the ranker only learns a regression in a fixed feature space. The finetuned 8B model is ahead on all nine datasets, by $\Delta = +0.019$ ROUGE-L on short-form generation and below $+0.01$ on long-form QA and explainable recommendation; measured against the headroom, the 30M ranker recovers 45\% to 94\% of the finetuned model's gain per dataset. This gap is the price of decoupling. The 8B model re-encodes every candidate and must be finetuned per dataset, whereas the ranker scores recycled embeddings at four orders of magnitude lower cost per query (Appendix~\ref{app:cost}); for that cost it delivers the bulk of what finetuning a billion-parameter model buys.

\begin{table}[t]
\centering
\caption{Agreement between ROUGE-L and BLEU at the candidate level, computed on the evaluation prompts of each dataset. Within-pool statistics correlate the two metrics over the 64 candidates of a prompt and are averaged over prompts; ``same argmax'' is the fraction of pools in which both metrics select the same best candidate; the across-prompt column correlates the per-prompt mean scores.}
\label{tab:metric_agreement}
\resizebox{\textwidth}{!}{%
\begin{tabular}{l ccc ccc ccc}
\toprule
 & \multicolumn{3}{c}{Short-form Generation} & \multicolumn{3}{c}{Long-form QA} & \multicolumn{3}{c}{Explainable Recommendation} \\
\cmidrule(lr){2-4}\cmidrule(lr){5-7}\cmidrule(lr){8-10}
Statistic & News & Scholarly & Tweet & Art & Lifestyle & Society & Amazon & Yelp & Google \\
\midrule
Within-pool Spearman & 0.548 & 0.545 & 0.578 & 0.385 & 0.405 & 0.410 & 0.594 & 0.389 & 0.363 \\
Within-pool Pearson & 0.529 & 0.520 & 0.614 & 0.394 & 0.418 & 0.434 & 0.638 & 0.404 & 0.383 \\
Same argmax & 31\% & 31\% & 37\% & 17\% & 12\% & 18\% & 28\% & 17\% & 19\% \\
Across-prompt Pearson & 0.778 & 0.496 & 0.647 & 0.564 & 0.484 & 0.699 & 0.743 & 0.685 & 0.688 \\
\bottomrule
\end{tabular}}
\end{table}

\paragraph{Cross-Metric Generalization.}
Because the ranker is trained on ROUGE-L targets, one may ask whether it merely regresses the training metric. If so, its advantage should vanish under a metric it never saw. Table~\ref{tab:bleu_transfer} therefore evaluates the Best-of-64 selection of every ranker under BLEU, keeping all rankers fixed; ours remains trained solely on ROUGE-L labels. Our ranker selects the highest-BLEU candidate on all nine datasets, and the four generalist reward models trail it everywhere, most clearly on explainable recommendation, where our ranker improves on the strongest baseline by 20\% to 64\% relative.

This transfer is not trivial. Table~\ref{tab:metric_agreement} quantifies how differently ROUGE-L and BLEU order the candidates of a pool. The two metrics select the same best candidate in only 12--37\% of pools, and their within-pool rank correlation is 0.36--0.59, so a ranker that had memorized the ROUGE-L scoring function would gain little under BLEU. The across-prompt correlation is higher (0.48--0.78), which reflects shared difficulty between prompts rather than agreement on which candidate to pick, and it is the within-pool disagreement that matters for selection. The ranker has therefore learned which candidate comes closest to what the user would have written, a property that transfers across surface metrics. The pattern mirrors the main results: generalist reward models fall short not because the evaluation is $n$-gram based, but because they lack calibration for user-authored references, a conclusion now corroborated under a metric none of the rankers was tuned toward.

\paragraph{User History Influence.}
Since the user profile is the only source of personalization signal, we examine how the amount of available history affects the pipeline. Table~\ref{tab:history_stats} summarizes the number of historical interactions per user on the evaluation prompts of each dataset, together with the profile text that enters the generation prompt and $h_u$. History richness spans more than an order of magnitude, from a median of eight recorded interactions on XRec to 143 article--headline pairs on News, and is highly dispersed within datasets as well. The main results therefore aggregate over heterogeneous history regimes. Following each benchmark's protocol, only a retrieved subset (three BM25-retrieved items for LaMP, six for LaMP-QA) or the pre-written summary (XRec) enters the prompt and $h_u$, never the full history. There is no minimum history and no per-user adaptation step.

Table~\ref{tab:history_stratified} buckets evaluation users by their number of historical interactions and reports the Best-of-64 ROUGE-L of the default ranker within each bucket. Because the model is unchanged, differences across buckets reflect only the history available to the pipeline. Bucket support mirrors Table~\ref{tab:history_stats}: XRec users, with at most 22 recorded interactions, fall into the two lowest buckets, Tweet users concentrate below 25 interactions, and Scholarly users concentrate above. Two patterns emerge. First, performance generally rises with available history, most clearly on Tweet (0.410 to 0.478), Scholarly (0.369 to 0.410), and all three XRec datasets. The long-form QA columns are flatter, with no consistent direction, and News is the exception, dropping for users with 100 or more pairs. Second, degradation under sparsity is graceful rather than catastrophic, as the sparsest users still retain most of the benefit of selection. We therefore read the direction of the trend as robust and the ordering of adjacent buckets as within noise.

\begin{table*}[t]
\centering
\caption{BoN ROUGE-L ($N{=}64$) stratified by the amount of history each user actually has. The selector is the paper-default ranker, so differences across buckets reflect history available to the pipeline, not a change of model. $^{\dagger}$Buckets with fewer than 10 users, reported for completeness only.}
\label{tab:history_stratified}
\resizebox{\textwidth}{!}{%
\begin{tabular}{l ccc ccc ccc}
\toprule
 & \multicolumn{3}{c}{\textbf{Short-form Generation}} & \multicolumn{3}{c}{\textbf{Long-form QA}} & \multicolumn{3}{c}{\textbf{Explainable Recommendation}} \\
\cmidrule(lr){2-4}\cmidrule(lr){5-7}\cmidrule(lr){8-10}
\textbf{History size} & News & Scholarly & Tweet & Art & Lifestyle & Society & Amazon & Yelp & Google \\
\midrule
$<$10 & 0.1498 & -- & 0.4104 & -- & -- & -- & 0.2640 & 0.1904 & 0.1675 \\
10--24 & 0.1832 & -- & 0.4091 & 0.1535 & 0.1470 & 0.1528 & 0.2659 & 0.1958 & 0.1729 \\
25--49 & 0.1954 & 0.3693 & 0.4197 & 0.1455$^{\dagger}$ & 0.1430 & 0.1440 & -- & -- & -- \\
50--99 & 0.1952 & 0.3900 & 0.4462 & 0.1477 & 0.1524 & 0.1556 & -- & -- & -- \\
100$+$ & 0.1385 & 0.4103 & 0.4781$^{\dagger}$ & 0.1620 & 0.1532 & 0.1407 & -- & -- & -- \\
\bottomrule
\end{tabular}}
\end{table*}

%% file: conclusion.tex
\section{Conclusion}
\label{section:conclusion}
We challenged the assumption that personalized LLM alignment requires expensive train-time intervention. An oracle Best-of-$N$ analysis showed that the base generator already produces well-personalized candidates but generalist reward models fail to select them, which reframes personalized alignment as a candidate selection problem. To solve it at low cost, we introduced a million-parameter MLP ranker that scores candidates directly from the generator's own hidden states. Trained on the generator's own candidate pools labeled against user-authored references, it learns the fine-grained distinctions that personalized selection requires. The ranker outperforms generalist 8B reward models on all nine datasets and keeps improving as the pool grows where they stall. It approaches an 8B reward model finetuned per dataset at four orders of magnitude lower scoring latency, and it is competitive with personalized reward models without any per-user preference labels. It can also steer decoding directly, recovering part of the selection gain from a single generation. Since it attaches to any generator that exposes its hidden states, it offers a practical route to per-user alignment without retraining the generator. Our evaluation rests on lexical overlap with user-authored references, and a human study of style consistency is left for future work.

%% file: appendix.tex
\section{Full Per-Dataset Results}
\label{app:full_results}
This appendix expands the aggregated figures of Section~\ref{section:experiments} to all nine datasets.

\subsection{Personalization Headroom Analysis}
\label{appendix:headroom_analysis}

\begin{figure*}[t]
    \centering
    \includegraphics[width=\textwidth]{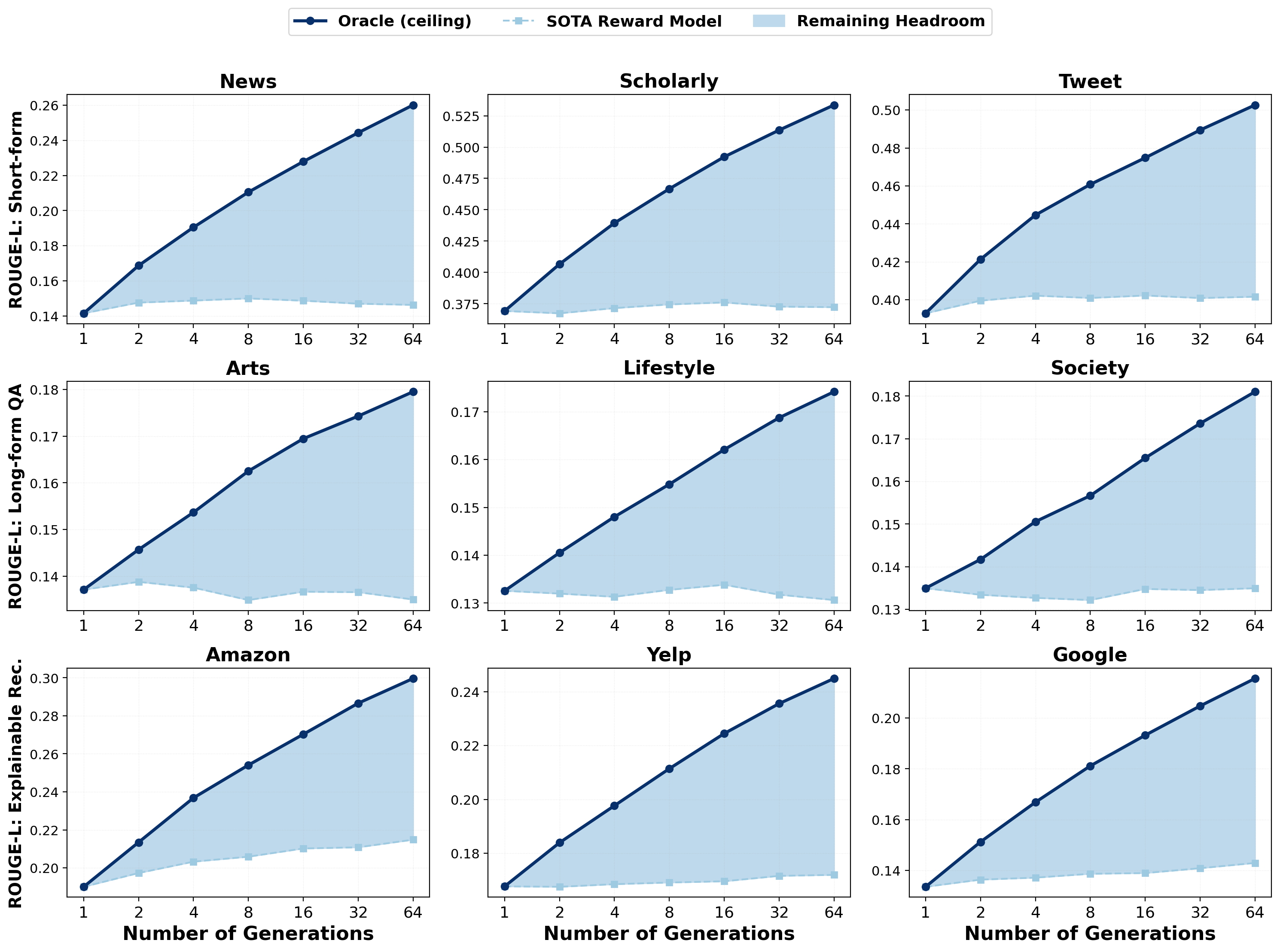}
    \caption{Personalization headroom under Best-of-$N$ sampling ($N \le 64$, ROUGE-L) on each of the nine datasets. The SOTA reward model line is the per-$N$ maximum over the four generalist reward models, and the shaded region is its gap to the Oracle ceiling.}
    \label{fig:headroom_full}
\end{figure*}

In Section~\ref{sec:headroom_analysis}, we presented the aggregated performance headroom across three distinct personalized generation tasks. To provide a more granular view of the test-time scaling behavior, Figure~\ref{fig:headroom_full} details the full Best-of-$N$ scaling trajectories for all nine individual datasets.

Across all nine datasets the pattern of Section~\ref{sec:headroom_analysis} holds. The Oracle ceiling (solid dark blue line) rises monotonically as the candidate pool $N$ expands from 1 to 64, whereas the strongest generalized reward model (dashed light blue line; the per-dataset maximum over the four reward models of Section~\ref{subsec:settings}) plateaus or declines. At $N=64$ it captures 4\%, 2\%, and 8\% of the oracle headroom on News, Scholarly, and Tweet and 23\%, 6\%, and 11\% on Amazon, Yelp, and Google, and $-5\%$, $-5\%$, and 0\% on Art, Lifestyle, and Society, where its Best-of-64 pick is no better than a single sample (ROUGE-L; on ROUGE-1 the shares range from $-9\%$ to 22\%). The base LLM thus produces well-matched candidates in every domain, and generalist reward models are miscalibrated for picking them out of a large pool.

\subsection{Baseline Comparisons Across Datasets}
\label{appendix:baseline_comparisons}

\begin{figure*}[t]
    \centering
    \includegraphics[width=\textwidth]{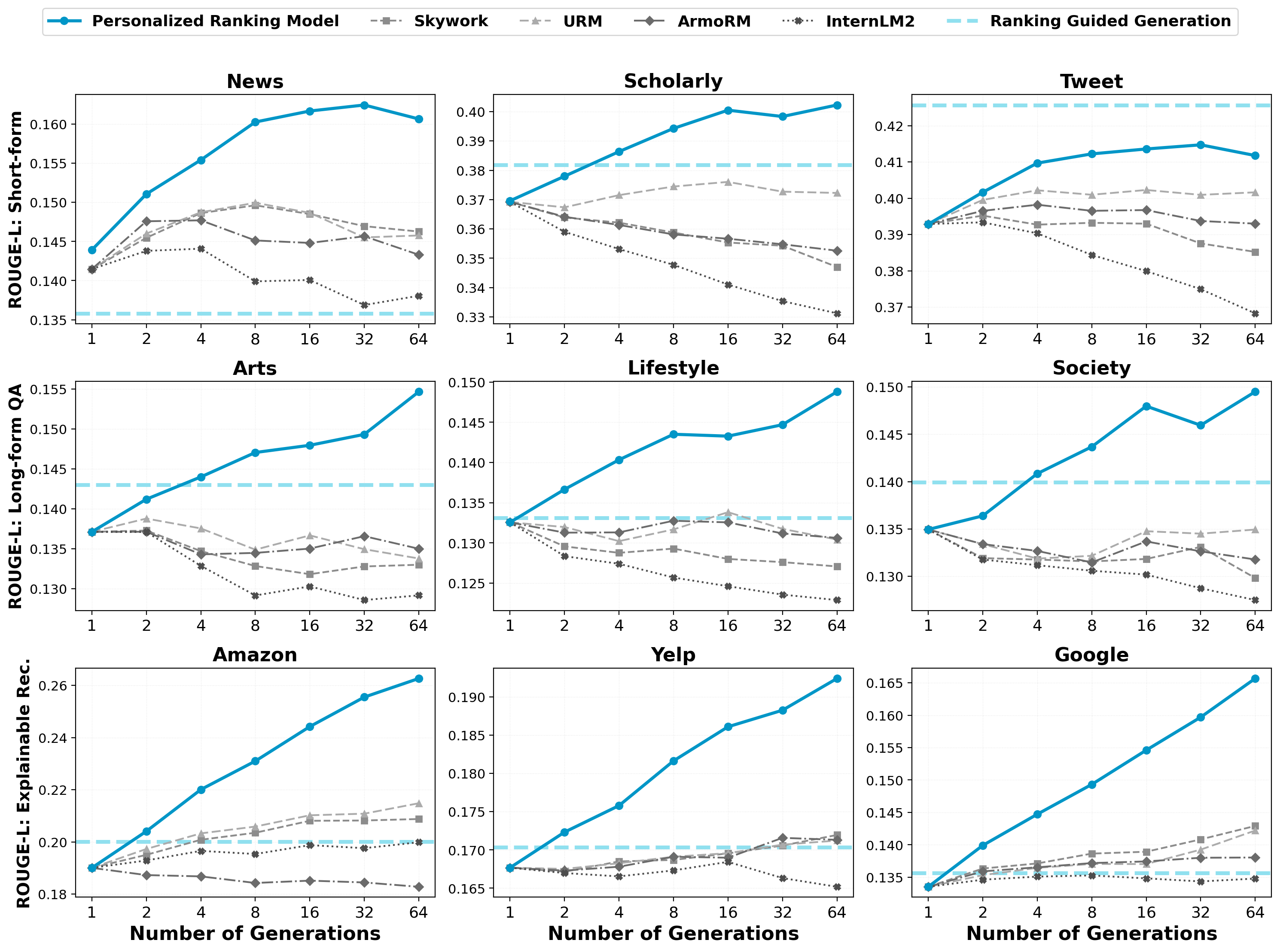}
    \caption{Best-of-$N$ selection (ROUGE-L) on each of the nine datasets with our Personalized Ranking Model and each of the four generalist reward models. The dashed horizontal line is ranking guided generation, one greedily decoded response per prompt (Appendix~\ref{app:rgg}).}
    \label{fig:rank_full}
\end{figure*}

Expanding upon the aggregated results presented in the main text, Figure~\ref{fig:rank_full} provides a comprehensive dataset-level breakdown of Best-of-$N$ scaling performance against individual state-of-the-art reward models (\texttt{Skywork-Reward-V2}, \texttt{URM-LLaMa-3.1}, \texttt{ArmoRM-LLaMa3}, and \texttt{InternLM2-Reward}).

\textbf{Consistent Degradation of General Reward Models.}
On the short-form datasets the reward models are flat on News (within 0.005 ROUGE-L of their $N=1$ score) and decline on Scholarly and Tweet, where \texttt{InternLM2-Reward} loses 0.038 and 0.025 between $N=1$ and $N=64$ and \texttt{Skywork-Reward-V2} loses 0.022 on Scholarly; only \texttt{URM} gains, by at most 0.009. On the three long-form QA datasets all four models end at or below their $N=1$ score. Explainable recommendation is the one task on which generalist models benefit from a larger pool: \texttt{Skywork-Reward-V2} and \texttt{URM} gain 0.019 and 0.025 on Amazon and up to 0.009 on Yelp and Google, while \texttt{ArmoRM} and \texttt{InternLM2-Reward} stay within 0.01 of $N=1$. Even there the best reward model captures no more than 23\% of the oracle headroom.

\textbf{Superiority of the Personalized Ranker and Guided Generation.}
Our Personalized Ranking Model (solid blue line) is the best selector at $N=64$ on all nine datasets, ahead of the best reward model by 0.010 (Tweet) to 0.048 (Amazon) ROUGE-L, and it improves with $N$ on every dataset; on News, Scholarly, Tweet, and Society the curve saturates around $N=16$ to $32$ and fluctuates by up to 0.003 thereafter. The dashed horizontal line is ranking guided generation, one greedily decoded response per prompt (Appendix~\ref{app:rgg}). It lies between Best-of-1 and Best-of-4 of our ranker on seven datasets, above Best-of-64 on Tweet (0.426 against 0.412) and below Best-of-1 on News (0.136 against 0.144), so steering a single trajectory recovers only a small part of the selection gain.

\subsection{Scaling Analysis and Parameter Efficiency}
\label{appendix:scaling_analysis}

\begin{figure*}[t]
    \centering
    \includegraphics[width=\textwidth]{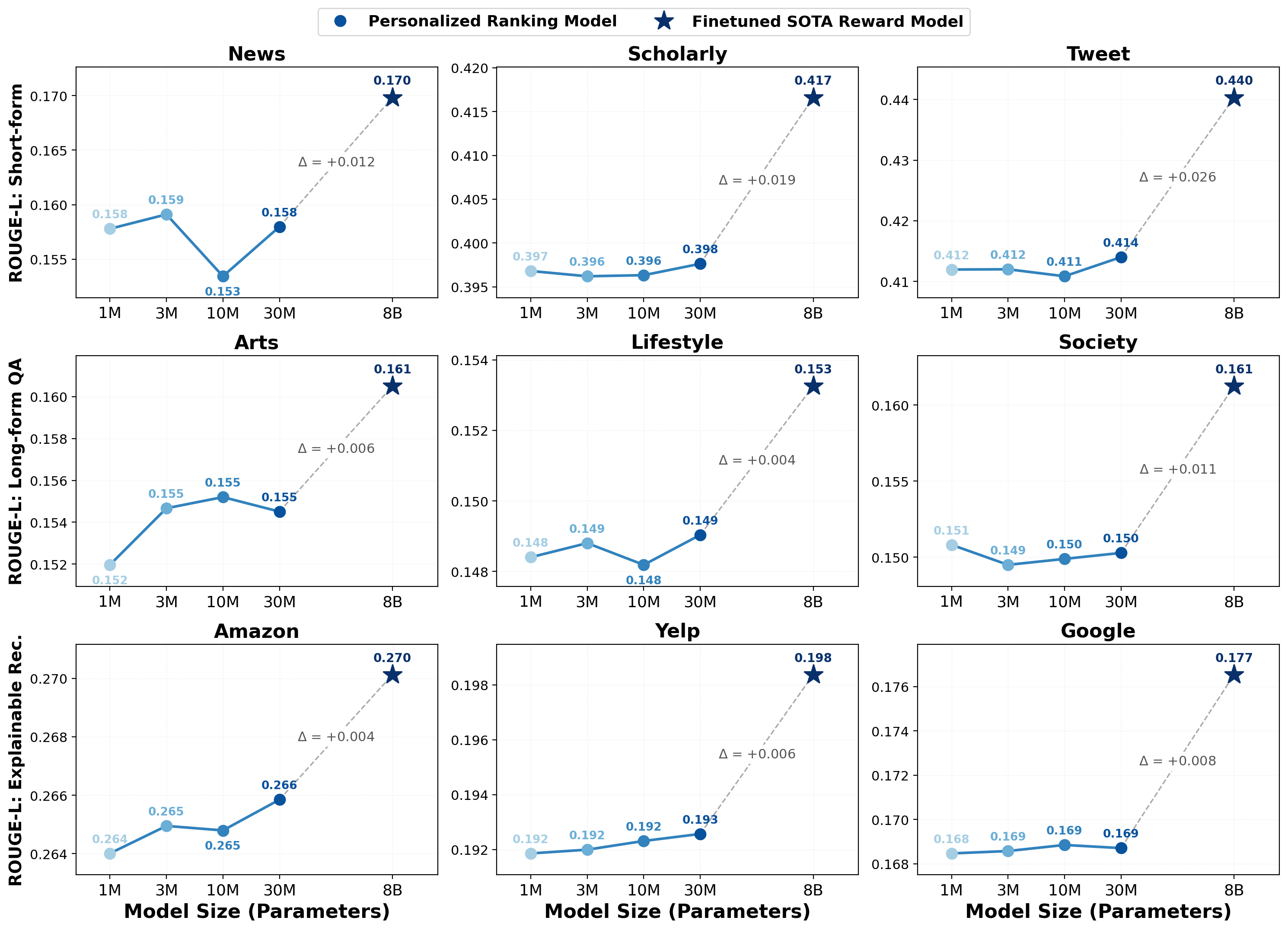}
    \caption{Ranker size against the finetuned 8B reward model at Best-of-64 (ROUGE-L) on each dataset. $\Delta$ is the 8B score minus the 30M score.}
    \label{fig:scale_full}
\end{figure*}

Figure~\ref{fig:scale_full} shows, for each dataset, the Best-of-64 ROUGE-L of our Personalized Ranking Model at 1M, 3M, 10M, and 30M parameters and of the finetuned 8B reward model of Appendix~\ref{app:training}. 

\textbf{Ranker size.} The ranker is insensitive to its size: across 1M to 30M parameters the score moves by at most 0.006 ROUGE-L (News) and by less than 0.002 on Scholarly, Lifestyle, Society, Amazon, Yelp, and Google, and the ordering of the four sizes is not consistent across datasets, so we treat the differences as noise and use the 2.8M model by default.

\textbf{Finetuned 8B reward model.} The finetuned model is ahead on every dataset: $\Delta$ (8B minus 30M) is $+0.012$, $+0.019$, and $+0.026$ on News, Scholarly, and Tweet, $+0.006$, $+0.004$, and $+0.011$ on Art, Lifestyle, and Society, and $+0.004$, $+0.006$, and $+0.008$ on Amazon, Yelp, and Google. Measured against the headroom between a single sample and the oracle, the 30M ranker captures 12\% to 66\% and the 8B model 22\% to 70\%, so the ranker recovers 45\% (Tweet) to 94\% (Amazon) of the finetuned model's gain at roughly $10^{-4}$ of its scoring latency and without finetuning an 8B model per dataset (Appendix~\ref{app:cost}).

\section{Additional Analyses}
\label{app:analyses}
This appendix reports the ablations and comparisons referenced in the main text: the training objective and personalized reward-model baselines.

\subsection{Ranking Objectives}
\label{sec:loss comparison}

Table~\ref{tab:loss_ablation} retrains the identical 2.8M-parameter MLP ranker under the pairwise and listwise objectives discussed in Section~\ref{subsec:loss function}, keeping the architecture, training data, and budget fixed (Appendix~\ref{app:training}). MSE regresses to the ROUGE-L target standardized within each candidate pool; Bradley--Terry is trained on sampled candidate pairs and RankNet on all pairs within a pool; ListNet, ListMLE, and LambdaRank consume the ordering of each pool induced by the targets. No single objective dominates: MSE is best on News, ListMLE on Tweet, Lifestyle, and Society, the pairwise objectives on Scholarly, Art, Yelp, and Google, and ListNet on Amazon. The spread across objectives is small, at most 0.012 ROUGE-L on any dataset and below 0.003 on the three explainable recommendation datasets. The objectives that fall furthest behind on the three short-form datasets and on Art and Lifestyle are ListNet and LambdaRank, which were designed for graded ordinal relevance in retrieval. We draw two conclusions. Pointwise MSE is a well-founded default in our setting, since it uses the magnitude information in the cardinal labels and scales linearly in the pool size, but pairwise objectives are not inherently suboptimal here and perform on par. More importantly, the narrow spread confirms that the gains of the framework stem from embedding recycling and train-time candidate scaling rather than from the loss, which we do not claim as a contribution.

\begin{table*}[t]
\centering
\caption{Ranking-objective ablation across all nine datasets. Identical MLP architecture, training data, and budget; only the training objective changes. Cells report BoN ROUGE-L at $N{=}64$. \textbf{Bold} marks the best objective per dataset.}
\label{tab:loss_ablation}
\resizebox{\textwidth}{!}{%
\begin{tabular}{l ccc ccc ccc}
\toprule
 & \multicolumn{3}{c}{\textbf{Short-form Generation}} & \multicolumn{3}{c}{\textbf{Long-form QA}} & \multicolumn{3}{c}{\textbf{Explainable Recommendation}} \\
\cmidrule(lr){2-4}\cmidrule(lr){5-7}\cmidrule(lr){8-10}
\textbf{Objective} & News & Scholarly & Tweet & Art & Lifestyle & Society & Amazon & Yelp & Google \\
\midrule
MSE & \textbf{0.1591} & 0.3962 & 0.4120 & 0.1547 & 0.1488 & 0.1495 & 0.2650 & 0.1920 & 0.1686 \\
Bradley--Terry & 0.1545 & 0.3980 & 0.4122 & \textbf{0.1558} & 0.1492 & 0.1512 & 0.2664 & 0.1928 & \textbf{0.1694} \\
RankNet & 0.1549 & \textbf{0.3981} & 0.4116 & 0.1557 & 0.1491 & 0.1515 & 0.2663 & \textbf{0.1931} & 0.1691 \\
ListNet & 0.1506 & 0.3936 & 0.4085 & 0.1502 & 0.1482 & 0.1526 & \textbf{0.2675} & 0.1923 & 0.1691 \\
ListMLE & 0.1551 & 0.3967 & \textbf{0.4135} & 0.1557 & \textbf{0.1492} & \textbf{0.1527} & 0.2670 & 0.1930 & 0.1691 \\
LambdaRank & 0.1483 & 0.3949 & 0.4021 & 0.1531 & 0.1465 & 0.1526 & 0.2661 & 0.1919 & 0.1687 \\
\bottomrule
\end{tabular}}
\end{table*}

\begin{table*}[t]
\centering
\caption{Cross-metric generalization: BLEU of the Best-of-$N$ ($N{=}64$) selected candidate across all nine datasets. Every ranker is fixed (ours remains trained solely on ROUGE-L labels); only the evaluation metric changes. \textbf{Bold} marks the best selector per dataset.}
\label{tab:bleu_transfer}
\resizebox{\textwidth}{!}{%
\begin{tabular}{l ccc ccc ccc}
\toprule
 & \multicolumn{3}{c}{\textbf{Short-form Generation}} & \multicolumn{3}{c}{\textbf{Long-form QA}} & \multicolumn{3}{c}{\textbf{Explainable Recommendation}} \\
\cmidrule(lr){2-4}\cmidrule(lr){5-7}\cmidrule(lr){8-10}
\textbf{Method} & News & Scholarly & Tweet & Art & Lifestyle & Society & Amazon & Yelp & Google \\
\midrule
\textbf{Ours} & \textbf{0.0387} & \textbf{0.0671} & \textbf{0.1150} & \textbf{0.0259} & \textbf{0.0217} & \textbf{0.0277} & \textbf{0.0822} & \textbf{0.0417} & \textbf{0.0337} \\
\textbf{Skywork} & 0.0330 & 0.0528 & 0.1050 & 0.0202 & 0.0166 & 0.0248 & 0.0498 & 0.0347 & 0.0274 \\
\textbf{InternLM2} & 0.0297 & 0.0503 & 0.0990 & 0.0205 & 0.0148 & 0.0235 & 0.0446 & 0.0332 & 0.0243 \\
\textbf{URM} & 0.0326 & 0.0593 & 0.1105 & 0.0214 & 0.0168 & 0.0251 & 0.0500 & 0.0347 & 0.0272 \\
\textbf{ArmoRM} & 0.0316 & 0.0544 & 0.1083 & 0.0225 & 0.0170 & 0.0247 & 0.0370 & 0.0347 & 0.0256 \\
\bottomrule
\end{tabular}}
\end{table*}

\subsection{Personalized Reward Model Baselines}
\label{app:personalized_rm}

Table~\ref{tab:personalized_rm} compares our ranker with reward models that personalize by fitting or inferring user-specific parameters from labeled preference comparisons of the target user: PAL~\citep{chen2024pal}, VPL~\citep{poddar2024personalizing}, PReF~\citep{shenfeld2025language}, LoRe~\citep{bose2025lore}, GPO~\citep{zhao2023group}, and SynthesizeMe~\citep{ryan2025synthesizeme}. We evaluate on XRec, the only family in which such supervision can be constructed: its users recur across interactions, whereas every LaMP and LaMP-QA prompt belongs to a distinct user with a history but no preference labels, so for those tasks the strongest feasible heavyweight comparator is the reward model finetuned on the personalized training distribution reported in Section~\ref{subsec:more analysis}.

\paragraph{Protocol.} We follow the official implementations and preserve the mechanism that defines each method, namely how its per-user parameters are obtained. For each evaluation user we hold out up to eight of their training interactions and mine labeled comparison pairs from the on-policy candidate pools of those interactions, with the chosen response being the candidate with higher ROUGE-L against that user's reference. PAL, PReF, and LoRe fit their per-user vectors on these pairs by gradient descent or regularized logistic regression, with their shared prototypes or reward bases trained on the training-user population; VPL encodes the same pairs with its pair encoder to infer its latent user embedding; GPO meta-learns across users and receives the user's labeled context interactions in context at test time; SynthesizeMe induces a persona from the same pairs and selects with a persona-prompted pairwise judge (Llama-3.1-8B-Instruct) run as a single-elimination tournament over the pool. PAL, VPL, PReF, and LoRe operate on the same frozen generator embeddings as our ranker at a matched parameter budget of about 2.8M, with the official default configurations: two prompt-conditioned prototypes with a cosine ideal-point score for PAL, sixteen basis reward features for PReF, a rank-8 basis for LoRe, and the official 128-dimensional, six-layer transformer for GPO; SynthesizeMe runs the released package with a persona search budget of five to ten candidates per user. All methods select from the same Best-of-64 candidate pools and are evaluated on each user's held-out test interaction. Our ranker receives only the content profile and zero preference labels.

\paragraph{Results.} Our ranker matches the strongest reward-factorization baseline, VPL, within noise on all three datasets, while requiring no annotated pairs per user and no per-candidate encoder pass. PAL, PReF, and LoRe are below ours on every dataset. GPO obtains the highest raw scores, by about 0.005 ROUGE-L per dataset, but still depends on per-user labeled examples and on post-hoc scoring of materialized candidates. SynthesizeMe selects close to random: its persona-prompted judge captures 1--2\% of the oracle headroom, and a generic judge without the persona is no worse (0.2033, 0.1677, and 0.1378 ROUGE-L on Amazon, Yelp, and Google), mirroring the miscalibration of generalist reward models in Section~\ref{subsec:main_results}. A further diagnostic explains why the user-conditioning machinery of these baselines does not pay off here: giving each user another user's fitted parameters or context set changes the selection by at most 0.0008 ROUGE-L for every trained method, including ours. Because every candidate is generated with the user's profile in the prompt, personalization is already realized at generation time, and what remains at selection is a calibrated judgment of which candidate comes closest to what the user would have written. This is precisely the judgment our lightweight ranker provides without any per-user labels, and it reinforces rather than competes with our central claim that the personalization headroom is a candidate matching problem.

\begin{table*}[t]
\centering
\caption{Comparison with personalized reward-model baselines on XRec (BoN ROUGE-L, $N{=}64$). All baselines are instantiated per user from that user's own labeled data, whereas ours uses only the content profile (zero preference labels). Ours is retrained under the seen-user protocol of this comparison, which is why it differs slightly from Table~\ref{tab:loss_ablation}.}
\label{tab:personalized_rm}
\begin{tabular}{l ccccccc}
\toprule
\textbf{Dataset} & Ours & VPL & PAL & PReF & LoRe & GPO & SynthesizeMe \\
\midrule
Amazon & 0.2641 & 0.2628 & 0.2596 & 0.2598 & 0.2503 & 0.2690 & 0.1962 \\
Yelp & 0.1906 & 0.1916 & 0.1902 & 0.1791 & 0.1798 & 0.1957 & 0.1687 \\
Google & 0.1685 & 0.1697 & 0.1614 & 0.1526 & 0.1544 & 0.1743 & 0.1381 \\
\bottomrule
\end{tabular}
\end{table*}

\begin{table}[t]
\centering
\caption{Historical interactions per user on the evaluation prompts of each dataset (median, mean, 10th and 90th percentiles, maximum), and the profile text that enters the generation prompt and $h_u$. XRec counts are lower bounds from the interactions present in our splits; the benchmark reports 18--25 interactions per user.}
\label{tab:history_stats}
\resizebox{\textwidth}{!}{%
\begin{tabular}{l l r r r r r l}
\toprule
\textbf{Dataset} & \textbf{History unit} & \textbf{Median} & \textbf{Mean} & \textbf{p10} & \textbf{p90} & \textbf{Max} & \textbf{Profile text entering the prompt and $h_u$} \\
\midrule
News & article--headline pairs & 143 & 172.4 & 11 & 442 & 634 & top-3 BM25-retrieved pairs \\
Scholarly & abstract--title pairs & 80 & 99.4 & 53 & 167 & 913 & top-3 BM25-retrieved pairs \\
Tweet & past tweets & 13 & 17.6 & 9 & 29 & 226 & top-3 BM25-retrieved tweets \\
Art & past questions & 73 & 159.1 & 14 & 370 & 991 & top-6 BM25-retrieved questions \\
Lifestyle & past questions & 36 & 111.6 & 12 & 262 & 1488 & top-6 BM25-retrieved questions \\
Society & past questions & 57 & 115.8 & 13 & 284 & 1488 & top-6 BM25-retrieved questions \\
Amazon & past reviews & 9 & 8.2 & 3 & 11 & 22 & pre-written user summary \\
Yelp & past reviews & 9 & 7.0 & 2 & 10 & 18 & pre-written user summary \\
Google & past reviews & 8 & 6.9 & 2 & 10 & 15 & pre-written user summary \\
\bottomrule
\end{tabular}}
\end{table}


\section{Implementation Details}
\label{app:impl}

\subsection{Datasets}
\label{app:datasets}
We evaluate on nine datasets spanning three personalized generation paradigms. In every case the input pairs a task query with the target user's interaction history, and the reference is text specific to that user, so success requires matching individual style rather than generic quality. Table~\ref{tab:history_stats} lists, for each dataset, the profile text that enters the generation prompt and $h_u$, and the number of evaluation prompts is given below.

\paragraph{Short-form personalized generation (LaMP).} Given a document and the user's history of past writings, the model produces a short user-styled text: a news headline for an article (News, 1,889 evaluation prompts), a title for a research-paper abstract (Scholarly, 2,498), or a rephrased tweet in the user's voice (Tweet, 1,495). Following the benchmark's retrieval-augmented protocol, the prompt carries the three profile items retrieved by BM25 for the current input, and the reference is the headline, title, or tweet the user actually wrote.

\paragraph{Long-form personalized QA (LaMP-QA).} Given an open-ended information-seeking question and the asker's narrative history, the model generates a long-form answer tailored to the asker's background and needs, across three topical domains: Arts \& Entertainment (Art), Lifestyle \& Personal Development (Lifestyle), and Society \& Culture (Society). The prompt follows the benchmark's retrieval-augmented template with the six past questions retrieved by BM25. LaMP-QA ships rubrics rather than reference answers; we use, for each question, the per-user reference answer released with Personalized RewardBench~\citep{ma2026personalized}, whose question identifiers coincide with the LaMP-QA test set, and split the questions 90/10 into ranker training and evaluation (77, 99, and 108 evaluation prompts).

\paragraph{Personalized explainable recommendation (XRec).} Given a user--item interaction together with the user profile and the item profile, the model generates a natural-language explanation of why the item suits the user, on three platforms (Amazon, Yelp, Google; 3,000 evaluation prompts each), with the user's own review-style explanation as reference. We use the benchmark's explainer prompt with its textual user and item summaries and generate with the plain instruction-tuned LLM, without the collaborative embedding tokens specific to XRec's own model.

Following the original benchmarks, we report ROUGE-L against these references (ROUGE-1 where noted), and BLEU as a held-out metric in Section~\ref{subsec:more analysis}.

\subsection{Training Details}
\label{app:training}
\paragraph{Candidate pools and labels.} \textit{Qwen2.5-7B-Instruct} generates 256 samples per prompt at temperature 1.0 with vLLM; Best-of-$N$ uses the first $N$ candidates in stored order, and $N=64$ throughout the appendix. Each candidate is labeled with its ROUGE-L against the reference, standardized within its pool for the pointwise objective (Section~\ref{subsec:loss function}).

\paragraph{Embeddings.} The three embeddings are final-layer last-token hidden states of the same frozen \textit{Qwen2.5-7B-Instruct} ($d = 3584$), taken over the task query, the serialized profile text of Table~\ref{tab:history_stats}, and each candidate. In our implementation they are obtained with vLLM's last-token pooling in a pass separate from generation, which is the cost we report as ``embedding pass'' in Table~\ref{tab:compute_cost}; a serving stack that exposes the generator's hidden states removes this pass entirely, since the states are produced during generation anyway. No embedding is fine-tuned.

\paragraph{Ranker.} The default ranker concatenates $(h_x, h_u, h_y)$ and applies three linear layers with hidden width 256, GELU activations, and dropout 0.1, for 2.8M parameters. The main-text protocol is given in Section~\ref{subsec:settings}. The appendix ablations (Appendices~\ref{sec:loss comparison} and~\ref{app:rgg}) share one trainer: AdamW with learning rate $5\times10^{-4}$ and weight decay 0.01 for 15 epochs at batch size 64, a single seed, and the same candidate pools, so that all rows of a table are directly comparable.

\paragraph{Finetuned reward-model comparator.} The 8B comparator of Section~\ref{subsec:more analysis} is \textit{Skywork-Reward-V2-Llama-3.1-8B}, fully finetuned per dataset (bf16, fully sharded data parallel on two NVIDIA B200 GPUs) with the same pointwise objective as our ranker: mean-squared error to the within-pool standardized ROUGE-L, with scores centred within each pool. Each training prompt contributes eight candidates from its pool (the best, the worst, and six sampled at random); we use 1,500 training prompts (all 690 to 966 available prompts for LaMP-QA), 24 prompts per optimizer step, one epoch (two for LaMP-QA), learning rate $2\times10^{-5}$ chosen on a News pilot against $5\times10^{-6}$, 10\% warmup, weight decay 0.01, gradient clipping at 1.0, sequences truncated to 2,048 tokens, and a single seed. The finetuned model then scores the first 64 candidates of every evaluation prompt.

\subsection{Ranking Guided Generation Hyperparameters}
\label{app:rgg}
\paragraph{Selection.} The decoding hyperparameters of Section~\ref{sec:ranking_guided} were set a priori from interpretable considerations rather than by grid search: $\tau$ controls how often the ranker intervenes, $H_{\max}$ and $\alpha_0$ control the strength of an intervention, and the warmup protects the opening tokens of the response, before the partial-answer state carries content. The same values ($\tau=1$, $H_{\max}=8$, $\alpha_0=5$, 3-token warmup) are used for all nine datasets.

\paragraph{Sensitivity.} Table~\ref{tab:rgg_sensitivity} varies each hyperparameter around its default while holding the others fixed, with greedy decoding and at most 48 new tokens. Performance is stable across the full grid: the max--min spread over all settings is at most 0.0052 ROUGE-L on every dataset (mean 0.0037). The threshold $\tau$ is the one knob with a visible effect on cost: the gate $H_t > \tau$ fires on 14--59\% of decoding steps at $\tau=0.5$, on 4--39\% at the default, and on at most 11\% at $\tau=2$, so $\tau$ offers practitioners a single interpretable compute--intervention trade-off. $H_{\max}$, $\alpha_0$, and the warmup length change the outcome by at most 0.005 ROUGE-L. Mean next-token entropy under greedy decoding is 0.19--0.43 nats on short-form generation, 0.46--0.52 nats on explainable recommendation, and 0.83--0.92 nats on long-form QA, which is why the gate fires far more often on the latter.

\begin{table*}[t]
\centering
\caption{Sensitivity of ranking guided generation to its decoding hyperparameters. Each block varies one hyperparameter while holding the others at their defaults ($\tau=1$, $H_{\max}=8$, $\alpha_0=5$, warmup 3). Cells report ROUGE-L with greedy decoding.}
\label{tab:rgg_sensitivity}
\resizebox{\textwidth}{!}{%
\begin{tabular}{l ccc ccc ccc}
\toprule
 & \multicolumn{3}{c}{Short-form Generation} & \multicolumn{3}{c}{Long-form QA} & \multicolumn{3}{c}{Explainable Recommendation} \\
\cmidrule(lr){2-4}\cmidrule(lr){5-7}\cmidrule(lr){8-10}
Setting & News & Scholarly & Tweet & Art & Lifestyle & Society & Amazon & Yelp & Google \\
\midrule
$\tau = 0.5$ & 0.1356 & 0.3823 & 0.4272 & 0.1409 & 0.1348 & 0.1385 & 0.2002 & 0.1701 & 0.1330 \\
$\tau = 1.0$ (default) & 0.1358 & 0.3819 & 0.4257 & 0.1430 & 0.1331 & 0.1399 & 0.2001 & 0.1703 & 0.1357 \\
$\tau = 2.0$ & 0.1359 & 0.3816 & 0.4235 & 0.1423 & 0.1381 & 0.1397 & 0.2030 & 0.1714 & 0.1354 \\
$\tau = 4.0$ & 0.1378 & 0.3816 & 0.4229 & 0.1411 & 0.1364 & 0.1388 & 0.2044 & 0.1711 & 0.1354 \\
\midrule
$H_{\max} = 4$ & 0.1349 & 0.3819 & 0.4276 & 0.1404 & 0.1350 & 0.1361 & 0.2012 & 0.1711 & 0.1322 \\
$H_{\max} = 16$ & 0.1379 & 0.3819 & 0.4257 & 0.1412 & 0.1347 & 0.1401 & 0.2011 & 0.1697 & 0.1357 \\
\midrule
$\alpha_0 = 1$ & 0.1379 & 0.3819 & 0.4257 & 0.1406 & 0.1365 & 0.1406 & 0.2015 & 0.1710 & 0.1360 \\
$\alpha_0 = 2$ & 0.1379 & 0.3819 & 0.4257 & 0.1418 & 0.1350 & 0.1411 & 0.2014 & 0.1700 & 0.1357 \\
$\alpha_0 = 10$ & 0.1349 & 0.3819 & 0.4281 & 0.1392 & 0.1348 & 0.1363 & 0.2006 & 0.1702 & 0.1333 \\
\midrule
warmup $= 0$ & 0.1375 & 0.3829 & 0.4257 & 0.1432 & 0.1342 & 0.1398 & 0.2002 & 0.1703 & 0.1357 \\
warmup $= 8$ & 0.1376 & 0.3819 & 0.4240 & 0.1424 & 0.1332 & 0.1397 & 0.2015 & 0.1703 & 0.1359 \\
\bottomrule
\end{tabular}}
\end{table*}

\subsection{Inference Cost}
\label{app:cost}
\begin{table}[t]
\centering
\caption{Measured inference cost on one NVIDIA RTX A6000 for one query with $N=64$ candidates. Reward-model numbers are the mean over Skywork-Reward-V2, InternLM2, URM, and ArmoRM (range in the text); generation and embedding use Qwen2.5-7B-Instruct with vLLM.}
\label{tab:compute_cost}
\resizebox{\textwidth}{!}{%
\begin{tabular}{l r r r}
\toprule
\textbf{Stage} & \textbf{Per candidate} & \textbf{Per query ($N=64$)} & \textbf{Peak GPU memory} \\
\midrule
Candidate generation, short-form (Qwen2.5-7B-Instruct) & 10.2 ms & 0.65 s & $\approx$40 GiB \\
Embedding pass (only if states are not recycled from generation) & 2.08 ms & 0.13 s & 25--39 GiB \\
Ranker scoring, ours (2.8M MLP, recycled embeddings) & 0.0015 ms & 0.098 ms & 0.93 GiB \\
Reward-model scoring (8B, mean of four) & 37.1 ms & 2.38 s & 20.6 GiB \\
\bottomrule
\end{tabular}}
\end{table}

Table~\ref{tab:compute_cost} reports measured wall-clock and memory costs on a single NVIDIA RTX A6000 (46\,GiB) for one query with 64 candidates. Scoring the pool with the 2.8M-parameter ranker takes 0.098\,ms at 0.93\,GiB peak GPU memory, roughly 655k candidates per second, when the candidate embeddings are recycled from generation. Scoring the same pool with an 8B reward model takes 1.7--3.9\,s at 18--22\,GiB (mean 2.4\,s over Skywork-Reward-V2, InternLM2, URM, and ArmoRM), a latency gap of four orders of magnitude and a memory gap of about $22\times$. Our current implementation does not yet read the states off the generation pass but recomputes them in a separate pooling pass (Appendix~\ref{app:training}); counting that pass, end-to-end selection costs 2.1\,ms per candidate, still $18\times$ cheaper than reward-model scoring, and the recycled figure is the architectural ceiling. The one-off cost of embedding a user's profile is 7\,ms (XRec), 96\,ms (Scholarly), and 118\,ms (LaMP-QA) per user when computed standalone and zero when recycled. For context, generating the pool itself costs 10\,ms per candidate (0.65\,s per query) on short-form generation and roughly an order of magnitude more on long-form QA (about 11\,s per query), so generation rather than selection dominates the test-time budget, and scoring with an 8B reward model costs as much as generating the entire pool it evaluates. Ranking guided generation removes the need to materialize the pool at all: each intervention costs one forward and backward pass through the MLP plus one $V \times d$ matrix--vector product, which is negligible relative to a generator forward pass, and the gate fires on a minority of decoding steps (4--39\% at the default threshold, Appendix~\ref{app:rgg}). Generation and embedding use vLLM, whereas the reward models run with HF transformers at batch size 32 as released, so part of the latency gap reflects serving implementation; the parameter and memory ratios do not.

%% file: checklist.tex
\section*{NeurIPS Paper Checklist}
\begin{enumerate}
\item {\bf Claims}
    \item[] Question: Do the main claims made in the abstract and introduction accurately reflect the paper's contributions and scope?
    \item[] Answer: \answerYes{} 
    \item[] Justification: Abstract and introduction

\item {\bf Limitations}
    \item[] Question: Does the paper discuss the limitations of the work performed by the authors?
    \item[] Answer: \answerYes{} 
    \item[] Justification: Section 5

\item {\bf Theory assumptions and proofs}
    \item[] Question: For each theoretical result, does the paper provide the full set of assumptions and a complete (and correct) proof?
    \item[] Answer: \answerNA{} 
    \item[] Justification: \answerNA{}

    \item {\bf Experimental result reproducibility}
    \item[] Question: Does the paper fully disclose all the information needed to reproduce the main experimental results of the paper to the extent that it affects the main claims and/or conclusions of the paper (regardless of whether the code and data are provided or not)?
    \item[] Answer: \answerYes{} 
    \item[] Justification: Section 4

\item {\bf Open access to data and code}
    \item[] Question: Does the paper provide open access to the data and code, with sufficient instructions to faithfully reproduce the main experimental results, as described in supplemental material?
    \item[] Answer: \answerYes{} 
    \item[] Justification: after authorlist

\item {\bf Experimental setting/details}
    \item[] Question: Does the paper specify all the training and test details (e.g., data splits, hyperparameters, how they were chosen, type of optimizer) necessary to understand the results?
    \item[] Answer: \answerYes{} 
    \item[] Justification: Section 4

\item {\bf Experiment statistical significance}
    \item[] Question: Does the paper report error bars suitably and correctly defined or other appropriate information about the statistical significance of the experiments?
    \item[] Answer: \answerYes{} 
    \item[] Justification: Section 4

\item {\bf Experiments compute resources}
    \item[] Question: For each experiment, does the paper provide sufficient information on the computer resources (type of compute workers, memory, time of execution) needed to reproduce the experiments?
    \item[] Answer: \answerYes{} 
    \item[] Justification: Appendix C.2/C.4
    
\item {\bf Code of ethics}
    \item[] Question: Does the research conducted in the paper conform, in every respect, with the NeurIPS Code of Ethics \url{https://neurips.cc/public/EthicsGuidelines}?
    \item[] Answer: \answerYes{} 
    \item[] Justification: reviewed

\item {\bf Broader impacts}
    \item[] Question: Does the paper discuss both potential positive societal impacts and negative societal impacts of the work performed?
    \item[] Answer: \answerYes{} 
    \item[] Justification: Personalized generation is directly related to societal impacts.
    
\item {\bf Safeguards}
    \item[] Question: Does the paper describe safeguards that have been put in place for responsible release of data or models that have a high risk for misuse (e.g., pre-trained language models, image generators, or scraped datasets)?
    \item[] Answer: \answerNA{} 
    \item[] Justification: \answerNA{}

\item {\bf Licenses for existing assets}
    \item[] Question: Are the creators or original owners of assets (e.g., code, data, models), used in the paper, properly credited and are the license and terms of use explicitly mentioned and properly respected?
    \item[] Answer: \answerYes{} 
    \item[] Justification: Section 4.1

\item {\bf New assets}
    \item[] Question: Are new assets introduced in the paper well documented and is the documentation provided alongside the assets?
    \item[] Answer: \answerYes{} 
    \item[] Justification: We release our code repo.

\item {\bf Crowdsourcing and research with human subjects}
    \item[] Question: For crowdsourcing experiments and research with human subjects, does the paper include the full text of instructions given to participants and screenshots, if applicable, as well as details about compensation (if any)? 
    \item[] Answer: \answerNA{} 
    \item[] Justification: \answerNA{}

\item {\bf Institutional review board (IRB) approvals or equivalent for research with human subjects}
    \item[] Question: Does the paper describe potential risks incurred by study participants, whether such risks were disclosed to the subjects, and whether Institutional Review Board (IRB) approvals (or an equivalent approval/review based on the requirements of your country or institution) were obtained?
    \item[] Answer: \answerNA{} 
    \item[] Justification: \answerNA{}

\item {\bf Declaration of LLM usage}
    \item[] Question: Does the paper describe the usage of LLMs if it is an important, original, or non-standard component of the core methods in this research? Note that if the LLM is used only for writing, editing, or formatting purposes and does \emph{not} impact the core methodology, scientific rigor, or originality of the research, declaration is not required.
    \item[] Answer: \answerYes{} 
    \item[] Justification: Section 3

\end{enumerate}